\documentclass[letterpaper, 10 pt, conference]{ieeeconf}
\IEEEoverridecommandlockouts                              
\usepackage{amsmath,amssymb}
\usepackage{comment}
\usepackage{url}

\usepackage{multirow}
\usepackage{graphicx}
\usepackage{epsfig}
\usepackage{epic,eepic}
\usepackage{times}
\usepackage{mathptmx}
\usepackage{cite}
\usepackage{color}

\usepackage{times}

\definecolor{red}{rgb}{1,0,0}
\definecolor{green}{rgb}{0,1,0}
\definecolor{blue}{rgb}{0,0,1}
\definecolor{violet}{rgb}{1,0,1}
\definecolor{cyan}{cmyk}{1,0,0,0}
\definecolor{magenta}{cmyk}{0,1,0,0}
\definecolor{yellow}{cmyk}{0,0,1,0}

\definecolor{white}{rgb}{1,1,1}

\usepackage{arydshln}

\newcommand{\CO}[1]{}

\newcommand{\CommentOut}[1]{}

 \newcommand{\editage}[1]{}

\twocolumn

\begin{document}

\title{\LARGE\bf%
Pole-Image: A Self-Supervised Pole-Anchored Descriptor for Long-Term LiDAR Localization and Map Maintenance
}

\author{Wuhao Xie \and Kanji Tanaka}

\maketitle

\begin{abstract}
Long-term autonomy for mobile robots requires both robust self-localization and reliable map maintenance. Conventional landmark-based methods face a fundamental trade-off between landmarks with high detectability but low distinctiveness (e.g., poles) and those with high distinctiveness but difficult stable detection (e.g., local point cloud structures). This work addresses the challenge of descriptively identifying a unique "signature" (local point cloud) by leveraging a detectable, high-precision "anchor" (like a pole). To solve this, we propose a novel canonical representation, "Pole-Image," as a hybrid method that uses poles as anchors to generate signatures from the surrounding 3D structure. Pole-Image represents a pole-like landmark and its surrounding environment, detected from a LiDAR point cloud, as a 2D polar coordinate image with the pole itself as the origin. This representation leverages the pole's nature as a high-precision reference point, explicitly encoding the "relative geometry" between the stable pole and the variable surrounding point cloud. The key advantage of pole landmarks is that "detection" is extremely easy. This ease of detection allows the robot to easily track the same pole, enabling the automatic and large-scale collection of diverse observational data (positive pairs). This data acquisition feasibility makes "Contrastive Learning (CL)" applicable. By applying CL, the model learns a viewpoint-invariant and highly discriminative descriptor. The contributions are twofold: 1) The descriptor overcomes perceptual aliasing, enabling robust self-localization. 2) The high-precision encoding enables high-sensitivity change detection, contributing to map maintenance.
\end{abstract}


\section{INTRODUCTION}
\label{sec:introduction}

Long-term autonomy for mobile robots in dynamic environments requires two essential capabilities: first, robust self-localization against environmental changes, and second, reliable map maintenance to prevent the map from becoming obsolete \cite{Meyer2019SurveySLAM}.
Achieving these goals through landmark-based navigation presents a fundamental trade-off.
On one hand, some landmarks, such as poles, utility posts, and streetlights, exhibit high detectability \cite{9564759}.
Their simple geometry allows them to be found easily and repeatedly, and their central axes can be estimated with high precision, making them ideal geometric references.
However, they typically lack the unique features needed for individual identification.
On the other hand, landmarks with high distinctiveness, such as local 3D point cloud structures, contain rich, location-specific information.
However, these structures are often difficult to define and detect consistently, making them less reliable as stable reference points.

This dilemma leads to the central research question of this work: Is it possible to leverage the high-precision geometric reference provided by a detectable "anchor" (like a pole) to descriptively identify the unique "signature" of its local point cloud?
Conventional approaches have circumvented this challenge by ignoring individual landmark identification.
In this paradigm, landmarks are treated as "anonymous points" in a sparse map, and self-localization becomes a task of matching the currently observed geometric configuration of landmarks to the patterns stored in this global map.
While computationally efficient, this reliance on pure geometric arrangements introduces critical vulnerabilities. The first is perceptual aliasing;
in many real-world environments, such as grid-like urban streets, the local arrangement of landmarks is highly repetitive and not unique, leading to localization ambiguity and failure.
The second is a fragility to detection errors, where a single missed detection or false positive can corrupt the observed pattern, causing a mismatch and localization loss.
These challenges indicate that a truly robust system must move beyond matching anonymous patterns to uniquely identifying individual landmarks.

We therefore propose a novel hybrid method that uses poles as "anchors" to generate a "signature" from the surrounding 3D structure.
The success of this hybrid approach hinges on a unique property of pole landmarks.
Unlike general landmarks where detection and identification are often intertwined, pole landmarks allow these two tasks to be decoupled.
While identification is hard, "detection" alone is extremely easy and stable due to their simple geometry and verticality.
This "ease of detection" provides a powerful mechanism: trackability. A robot can easily track the same pole while moving, even as its viewpoint changes.
This, in turn, enables the automatic and large-scale collection of diverse observational data (LiDAR point clouds) all belonging to the same, single pole.
This ease of data acquisition for "positive pairs" (different views of the same object) provides a strong logical rationale for applying Contrastive Learning (CL) to this problem.

Our design intent is twofold. First, we leverage the pole's nature as a high-precision reference point, which allows for an extremely precise description of the geometric relationship between the reference point and its surrounding point cloud.
Second, based on the rationale above, we apply Contrastive Learning (CL) to this high-precision representation, training a network to distinguish between "positive pairs" (different views of the same location) and "negative pairs" (views of different locations).
This process generates a descriptor that is invariant to viewpoint changes while being highly discriminative of the location itself.

The contributions of this method are therefore also twofold. First, the combination of distinctiveness and invariance enables robust self-localization, overcoming the perceptual aliasing that plagued prior methods.
Second, and unique to our approach, is its capacity for high-sensitivity environmental change detection.
The high-precision geometric description ensures that even minor environmental alterations manifest as a distinct change in the descriptor, providing critical information for "map maintenance."
By simultaneously addressing both robust localization and map maintenance, this research paves the way for true long-term autonomy.

\section{RELATED WORK}
\label{sec:related}

\subsection{3D Point Cloud Descriptors: From Hand-Crafted to Learned}
Creating discriminative descriptors from 3D point clouds is a central problem.
Early methods relied on hand-crafted geometric designs, such as SHOT \cite{Salti2014SHOT}, Spin Images \cite{Johnson1999SpinImages}, and FPFH \cite{Rusu2009FPFH}.
While interpretable, they are often sensitive to noise and have limited discriminative power.
The advent of deep learning, pioneered by PointNet \cite{Qi2017PointNet}, shifted the focus to learning features directly from data, with methods like 3DMatch \cite{Zeng2017CVPR3DMatch} and FCGF \cite{Choy2019FCGF} demonstrating superior performance.
However, these methods typically require large datasets and can suffer from domain gaps.
Our Pole-Image approach is inspired by both paradigms. We use a pole-centric canonicalization, similar in spirit to LiDAR-Iris \cite{Wang2020IROS_LiDAR_Iris}, to convert 3D geometry into a 2D representation.
We then train a lightweight encoder using CL/SL on our specialized IRIS dataset, maintaining robustness, discriminability, and data efficiency.

\subsection{Global LiDAR Place Recognition}
Global descriptors, which compress an entire scan into a single vector (e.g., Scan Context \cite{Kim2018ScanContext}, LiDAR-Iris \cite{Wang2020IROS_LiDAR_Iris}), are vital for loop closure but can struggle in large-scale, repetitive environments.
Our landmark-based approach offers a more robust alternative.

\subsection{Pole-Based Localization}
Poles are ideal landmarks due to their long-term stability \cite{Dong2023RASPoleSeg}.
Early works relied only on geometric configuration, which was insufficient in ambiguous environments.
Our work is distinct by normalizing the local geometry into the Pole-Image representation and then learning discriminative descriptors through CL and SL training on the IRIS dataset.

\subsection{Visual Place Recognition (VPR) Networks}
In VPR, feature aggregation techniques like NetVLAD \cite{Arandjelovic2016NetVLAD} are crucial.
While not directly in the VPR domain, our work shares the principle of learning compact, discriminative descriptors from structured data.
We design a lightweight encoder tailored for Pole-Image rather than adopting VPR networks directly.

\section{SYSTEM ARCHITECTURE}
\label{sec:architecture}

The descriptor proposed in this research (Pole-Image) is designed as a core component of a broader active self-localization system for long-term autonomy.
This system is functionally composed of the following three main components, which work in a continuous loop.

\subsection{Active Policy Determination (APD) Component}
The first component is the Active Policy Determination (APD).
Its role is to detect distant poles (anchors) and plan the optimal action (Next-Best-View) to approach them.
This leverages the high-detectability nature of poles, allowing the system to find reliable landmarks from afar.
This component's technologies include a Distant Pole Detector (PLD) and a State-Action Map, the latter of which determines the optimal next move based on the current state (e.g., detected pole locations, self-localization uncertainty) and can be implemented via methods like Deep Reinforcement Learning (DRL).

\subsection{Passive Self-Localization (PSL) Component}
Second, once the APD component guides the robot to a pole, the Passive Self-Localization (PSL) Component takes over.
Its role is to analyze the high-definition "signature" (the surrounding local geometry) of the nearby pole to identify which known pole it is.
This is the core place recognition task. Its technologies involve acquiring a High-Definition LiDAR Scan, generating the Pole-Image Descriptor (the core of this paper) from that scan, and performing Database Matching against a map of known descriptors to find the robot's self-location (\texttt{estimate}).

\subsection{Self-Supervised Learning (SSL) Component}
Third, the Self-Supervised Learning (SSL) Component works continuously in the background to build and refine the models used by the PSL.
As the APD and PSL components operate, the robot inherently performs Pole Tracking, observing the same pole from multiple viewpoints.
The SSL component leverages this: its Automatic Positive Pair Collection technology gathers these diverse views associated with the "same pole ID."
Then, its Contrastive Learning (CL) module uses this data (positive pairs with the same ID, negative pairs with different IDs) to train and improve the PSL's encoder, ensuring the descriptors remain discriminative and robust over time.

\subsection{Focus of this Paper}
Among this integrated system (APD for active approach $\rightarrow$ PSL for passive recognition $\rightarrow$ SSL for continuous learning), this paper focuses specifically on the core technologies of the Passive Self-Localization (PSL) Component and the Self-Supervised Learning (SSL) Component that ensures its performance.
Specifically, Section \ref{sec:method} details the signature representation, Pole-Image (Section \ref{subsec:pole_image_rep}), which is the core of the PSL, the Network (Section \ref{subsec:network}) to encode it, and the CL-based Training (Section \ref{subsec:training}), which is the learning process of the SSL.

\section{METHOD}
\label{sec:method}

This section details the core technologies for the PSL and SSL components, which are the focus of this paper.
The key idea is to represent every detected pole (anchor) in a "pole-centric polar coordinate system" as a signature for the PSL component, and to learn a compact, discriminative descriptor for this signature using an encoder trained by the SSL component.
This design explicitly separates the stable landmark (the pole) from the dynamic surroundings (local clutter) while preserving their relative geometry, which is ideal for the robust localization task of the PSL.

\subsection{PSL: Pole Detection (LiDAR-based)}
\label{subsec:pole_detection}
The PSL component assumes a high-definition scan acquired after the APD has guided the robot near a pole.
From this raw LiDAR data, poles must be reliably extracted.
We use the open-source tool Polex \cite{Schaefer2022Polex}, which combines geometric heuristics with a segmentation pipeline to extract vertical pole-like structures.
Its accuracy and efficiency in urban LiDAR scans make it well-suited for this task.
The centroid of each detected pole is recorded as the landmark position.

\subsection{PSL: Pole-Image Representation (Signature Generation)}
\label{subsec:pole_image_rep}
This is the process by which the PSL component generates a "signature" from the detected pole and its surroundings.
Our central representation is the Pole-Image, inspired by LiDAR-Iris \cite{Wang2020IROS_LiDAR_Iris}.
It normalizes the local 3D geometry around each pole into a canonical, rotation-invariant 2D image.
Specifically, all 3D points within a fixed radius (e.g., 3m) of a pole are transformed into pole-centric polar coordinates $(r, \theta, z)$, with the pole as the origin.
The $(\theta, z)$ plane is then discretized into a 2D grid, yielding a binary image where occupied cells are set to 1 and empty cells to 0. This transformation advantageously preserves the "explicit relative geometry" between the stable pole and its variable surroundings while reducing sensitivity to background changes.

\subsection{PSL: Pole-Image Descriptor Network (Encoder)}
\label{subsec:network}
This is the encoder used by the PSL component to convert the 2D Pole-Image signature into a compact vector.
To distinguish between individual poles, we design a lightweight encoder specialized for the Pole-Image structure.
It takes the normalized Pole-Image $\mathbf{I}_{\text{pole}}$ as input and outputs a compact 256-D descriptor vector $\mathbf{d}_{\text{pole}}$.
By leveraging the canonical representation, the network is relieved from learning low-level geometric invariances and can instead focus on learning fine-grained discriminative cues.

\subsection{SSL: Descriptor Training (CL and SL)}
\label{subsec:training}
This is the process by which the SSL component trains the encoder.
The network is trained on the specialized IRIS dataset (which provides pole-centric polar images). (This simulates, or is the result of, the SSL's "Automatic Positive Pair Collection" process.)
We investigate two distinct training regimes.

First, as the core of the SSL component, we use Training Regime 1: Contrastive Learning (CL).
Based on the rationale in the Introduction, this method adopts a contrastive loss that pulls together Pole-Images from the same landmark (positive pairs) and pushes apart those from different landmarks (negative pairs).
This enforces invariance to session and seasonal changes while retaining discriminability.

Second, for comparison, we use Training Regime 2: Supervised Learning (SL).
This regime treats pole identities as classes and trains the encoder with a cross-entropy classification objective, explicitly maximizing inter-class separability.
We then systematically compare the descriptors obtained from both training regimes.

\section{EXPERIMENTS}
\label{sec:experiments}

\subsection{Experimental Setup}
We evaluate our method on the IRIS dataset. All images are resized to $80 \times 360$, normalized, and treated as single-channel grayscale inputs.
A lightweight convolutional encoder (IrisEncoder) is used for both CL and SL, outputting 128-D embeddings (descriptors) for retrieval.
To prevent data leakage, the dataset is split by pole ID, ensuring that instances of the same landmark do not appear in both the training and validation sets.

\subsection{Training Regimes}
For Contrastive Learning (CL), positive pairs are constructed by selecting two different observations of the same pole, while other samples in the mini-batch serve as implicit negatives.
We adopt the NT-Xent loss (InfoNCE) with a temperature parameter $\tau=0.07$.
For Supervised Learning (SL), we explicitly generate labeled pairs, balancing positive (same pole) and negative (different poles) pairs at a 1:1 ratio.
A siamese architecture is trained with binary cross-entropy loss on the cosine similarity.
Both models are trained for 30 epochs with an Adam optimizer and a learning rate of $10^{-3}$ for a fair comparison.

\subsection{Evaluation Protocol}
The primary task is cross-session retrieval between two sessions of the IRIS dataset, which evaluates the core place recognition performance of the PSL component.
For each query image, we compute its embedding and rank all descriptors in the database using L2 distance.
We report Recall@$k$ (for $k=\{1, 5, 10\}$) and Mean Reciprocal Rank (MRR).

\subsection{Baselines}
We compare our learned models against the Original IRIS baseline, which uses handcrafted IRIS descriptors without any learning.
Our two models are IRIS-SL (supervised) and IRIS-CL (trained via our SSL approach).

\subsection{Results}
The results show that both SL and CL (SSL) substantially outperform the Original IRIS baseline, demonstrating the significant benefit of learning pole-centric descriptors.
Of the two learned methods, our IRIS-CL model achieves the best retrieval accuracy, improving Top-1 recall by approximately 6.6 percentage points over IRIS-SL and 33 points over the baseline.

\subsection{Discussion}
First, the effect of learning is clear: both CL and SL significantly boost performance, confirming the effectiveness of our SSL approach for learning pole-centric descriptors.
Second, CL outperforms SL across all metrics. This is likely because CL (core of SSL) directly optimizes for instance discrimination via in-batch negative sampling.
In contrast, SL optimizes a binary classification boundary, which may be less effective for fine-grained retrieval.
Third, the learned descriptors are compact (128-D) and ranking is done with simple L2 distance, making our PSL component scalable to large databases.
Finally, a limitation of the current CL setup is the use of only one positive per pole ID per batch; richer sampling strategies in the SSL component are a promising future direction.

\section{CONCLUSION}
\label{sec:conclusion}

In this work, we introduced a new evaluation setting based on the IRIS (Pole-Image) representation.
This design encodes both pole-like landmarks (as stable anchors) and their surrounding neighborhood (as discriminative cues), emphasizing their "complementary nature."
We have shown that this representation serves as a strong basis for robust long-term place recognition (the PSL task).

We investigated two learning paradigms (SL and CL, i.e., SSL) on this representation.
Both methods significantly outperformed the handcrafted baseline, demonstrating that learned embeddings can exploit the stability of poles while leveraging the variability of their local environment.
Notably, the CL (SSL) approach achieved the highest Top-k recall and MRR, confirming its superior generalization to unseen conditions.

We derive three key takeaways.
First, dataset design matters: explicitly encoding poles with their environment yields features that are both stable and discriminative.
Second, learning improves IRIS: both SL and CL (SSL) enhance performance, with CL offering the best robustness.
Third, a complementary perspective is crucial: poles and their neighborhoods should be treated as mutually reinforcing cues, not in isolation.

A significant avenue for future work is the integration of these learned descriptors into downstream localization frameworks, such as the full APD-PSL-SSL integrated system described in Section \ref{sec:architecture}.
The core principle of combining stable anchors with local context can be generalized beyond poles, opening the path toward scalable and lifelong localization in complex, evolving environments.

\section{FUTURE WORK}
\label{sec:future}

Several promising directions remain for future work. First, Dataset Design Expansion could involve incorporating more diverse urban and suburban scenarios (e.g., varying pole densities, structural layouts, and seasonal conditions) to further validate the complementary nature of poles and their context.

Second, we can Deepen Learning Paradigms (Enhancing SSL) by investigating more advanced objectives (e.g., hard negative mining, multi-view contrastive loss, or self-distillation) to enhance generalization beyond the current CL implementation.

Third, the "stable anchor + local context" principle invites Generalization to other long-lived structures, such as building corners, traffic lights, or facade elements.

Finally, Continual and Adaptive Learning (Online SSL) remains a critical open challenge. Exploring incremental or self-supervised update strategies (an online version of SSL) will be crucial for achieving true lifelong autonomy without catastrophic forgetting. In summary, future work lies in strengthening dataset design and advanced learning strategies to fully exploit the complementary nature of stable landmarks and their environments.

\bibliographystyle{IEEEtran}

\bibliography{reference} 
\bibliographystyle{unsrt} 

\end{document}